\documentclass[conference]{IEEEtran}
\IEEEoverridecommandlockouts
\usepackage{cite}
\usepackage{booktabs}
\usepackage{amsmath,amssymb,amsfonts,amsthm}
\usepackage{algorithmic}
\usepackage{algorithm}
\usepackage{graphicx}
\usepackage{textcomp}
\usepackage{xcolor}
\def\BibTeX{{\rm B\kern-.05em{\sc i\kern-.025em b}\kern-.08em
    T\kern-.1667em\lower.7ex\hbox{E}\kern-.125emX}}
    
\theoremstyle{remark}

\begin{document}

\title{Undersmoothing Causal Estimators\\ with Generative Trees}

\author{\IEEEauthorblockN{Damian Machlanski}
\IEEEauthorblockA{
\textit{Department of Computer Science}\\
\textit{and Electronic Engineering}\\
\textit{University of Essex}\\
Colchester, UK \\
d.machlanski@essex.ac.uk}
\and
\IEEEauthorblockN{Spyros Samothrakis}
\IEEEauthorblockA{
\textit{Institute for Analytics and Data Science}\\
\textit{University of Essex}\\
Colchester, UK \\
ssamot@essex.ac.uk}
\and
\IEEEauthorblockN{Paul Clarke}
\IEEEauthorblockA{
\textit{Institute for Social and Economic Research}\\
\textit{University of Essex}\\
Colchester, UK \\
pclarke@essex.ac.uk}
}

\maketitle

\begin{abstract}
Inferring individualised treatment effects from observational data can unlock the potential for targeted interventions. It is, however, hard to infer these effects from observational data. One major problem that can arise is covariate shift where the data (outcome) conditional distribution remains the same but the covariate (input) distribution changes between the training and test set. In an observational data setting, this problem is materialised in control and treated units coming from different distributions. A common solution is to augment learning methods through reweighing schemes (e.g. propensity scores). These are needed due to model misspecification, but might hurt performance in the individual case. In this paper, we explore a novel generative tree based approach that tackles model misspecification directly, helping downstream estimators achieve better robustness. We show empirically that the choice of model class can indeed significantly affect the final performance and that reweighing methods can struggle in individualised effect estimation. Our proposed approach is competitive with reweighing methods on average treatment effects while performing significantly better on individualised treatment effects.
\end{abstract}

\section{Introduction}
In the absence of data from randomised experiments, analysts must use observational data to make inferences about the causal effects of interventions or treatments, that is, what would happen if they intervened to change the treatment status of individual units in a population. The estimation of average causal effects --- the average effect of the treatment aggregated across every unit in a population --- has been studied in considerable depth. However, there is now growing interest in estimating heterogeneous treatment effects for individuals characterized by a possibly large number of input variables or covariates. If there is substantial heterogeneity across units, such systems can unlock the analysis of targeted interventions, for instance, in the form of personalised healthcare based on covariates that describe patients' symptoms and health histories.

The use of observational data creates challenges for the estimation of heterogeneous causal effects.  First, the analyst must make assumptions, for example, that treatment selection is strongly ignorable given the available covariates. We take ignorability to hold throughout, and focus on the second problem, namely, that nonrandom treatment selection can lead to observed data in which the distributions of covariates among the treated and untreated units are very different. In practice, this can make it difficult for conventional learners to learn the true relationship between the treatment effect and covariates across the entire support of the covariates, and so result in poor performance when tested on other datasets.

More generally, this issue is known as `covariate shift', which in this setting means the learning target $P(Y|X)$ remains unchanged, while the marginal distributions of the covariate inputs $P(X)$ for treated and untreated can be very different. Most existing methods attempt to transform the observational distribution by sample reweighing schemes usually based on propensity scores \cite{chernozhukovDoubleDebiasedMachine2018, robinsEstimationRegressionCoefficients1994, rosenbaumCentralRolePropensity1983, kunzelMetalearnersEstimatingHeterogeneous2019, atheyGeneralizedRandomForests2019} (but not exclusively, see e.g.\ domain adaptation methods). However, reweighting seeks to standardise the observed support of $X$ for the treated and untreated groups, and so generally performs well for estimating treatment effects averaged across the common support of $X$, but less so for estimating conditional average treatment effects at points outside the observed support; in other words, as pointed out by \cite{wenRobustLearningUncertain2014}, reweighting does not address the problem of model misspecification which can be detrimental when it comes to estimating individualised treatment effects \cite{whiteConsequencesDetectionMisspecified1981}.

A promising alternative to these classical approaches is undersmoothing, where the model is allowed to fit the data very closely to capture $P(X)$ in the two groups, and in doing so potentially produce more accurate individualised predictions. Encouraged by suggestions elsewhere - \cite[footnote 3]{chernozhukovDoubleMachineLearning2016} and \cite{neweyUndersmoothingBiasCorrected1998} - in this paper, we develop a novel approach to causal effect estimation that improves accuracy by undersmoothing the observed data. 

Specifically, we propose to undersmooth using fast and straightforward generative trees \cite{correiaJointsRandomForests2020a} to augment the existing data, and in doing so facilitate more robust learning of downstream estimators of key causal parameters. The trees are used to `discretise' the input space into subpopulations of similar units (subclassification); the distributions of these groups are then modelled separately via mixtures of Gaussians, from which we sample equally to reduce data imbalances.

Data augmentation has proven effective in multiple scenarios. For instance, image transformations in computer vision \cite{perezEffectivenessDataAugmentation2017}, or oversampling minority classes in imbalanced classification problems \cite{chawlaSMOTESyntheticMinority2002, heADASYNAdaptiveSynthetic2008}. In our case, the method we propose could be seen as oversampling underrepresented data regions instead of just classes.

Generative models have also been investigated in causal inference literature \cite{atheyUsingWassersteinGenerative2020, nealRealCauseRealisticCausal2021}, though mostly for benchmarking purposes, where new synthetic data sets are created that closely resemble real data but with access to true effects. This work, on the other hand, goes beyond data modelling and focuses on targeted data augmentation instead.

Arguably the closest work to ours that combines data augmentation and generative models within the causal inference setting is \cite{bicaEstimatingEffectsContinuousvalued2020}. Despite a similar approach on a high-level, that is, train downstream causal estimators on augmented data, we believe our frameworks differ substantially upon further examination. More precisely, \cite{bicaEstimatingEffectsContinuousvalued2020} incorporates neural network based generative models to specifically generate counterfactuals and focuses on conditions where the treatment is continuous. In this work, our proposed method: a) is based on simple and widely-used decision trees, b) does not specifically generate counterfactuals, but oversamples heterogeneous data regions (more general), and c) works with classic discrete treatments.

In terms of this paper's contributions, we show empirically that the choice of model class can have a substantial effect on estimator's final performance, and that standard reweighing methods can struggle with individual treatment effect estimation. Given our experiments, we also provide an evidence that our proposed method increases data complexity that leads to statistically significant improvements in individual treatment effect estimation, while keeping the average effect predictions competitive. Our experimental setup incorporates a wide breadth of non-neural standard causal inference methods and data sets. We specifically focus on non-neural solutions as they are more commonly used by practitioners. The code accompanying this paper is available online\footnote{https://github.com/misoc-mml/undersmoothing-data-augmentation}.

The rest of the document is structured as follows. First, we revisit fundamental concepts that should aid understanding of the technical part of the paper. Next, we formally discuss the problem of model misspecification, followed by a thorough description of our proposed method. We then present our experimental setup and obtained results. Next section provides further discussion on the results, their implications and considered limitations of the method. Final section concludes the paper.

\section{Preliminaries}
This section gives a brief overview of the essential background deemed relevant to this work. For a more extensive review, we refer the reader to classic positions on causal analysis \cite{pearl2009causality, petersElementsCausalInference2017}, and recent surveys on causal inference \cite{guoSurveyLearningCausality2020, yaoSurveyCausalInference2020}.

Given two random variables $T$ and $Y$, investigating effects of interventions can be described as measuring how the outcome $Y$ differs across different inputs $T$. Real world systems usually contain other background covariates, denoted as $X$, which have to be accounted for in the analysis as well. To formally approach the task, we take Rubin's Potential Outcomes \cite{rubinEstimatingCausalEffects1974} perspective, which is particularly convenient in outcome estimation without knowing the full causal graph.

We start by defining the potential outcomes $\mathcal{Y}_t^{(i)}$, that is, the observed outcome when individual $i$ receives treatment $t=0,1$. Given this, the Individual Treatment Effect (ITE) can be written as:
\begin{equation}\label{eq:ite}
    ITE_i = \mathcal{Y}_1^{(i)} - \mathcal{Y}_0^{(i)}
\end{equation}
Thus, to compute such a value for individual $i$, we need access to both potential outcomes, $\mathcal{Y}_1^{(i)}$ and $\mathcal{Y}_0^{(i)}$, but only one, called the \textit{factual}, is observed:\ the other potential outcome, called the \textit{counterfactual}, cannot be observed. The fact that we only observe factuals but also need the counterfactuals to properly compute causal effects is known as the fundamental problem of causal inference:\ ITEs are not identified by the observed data.

However, parameters such as the Average Treatment Effect (ATE) and Conditional Average Treatment Effect (CATE) are identified, where
\begin{equation}\label{eq:ate}
    ATE = \mathbb{E}\left [ \mathcal{Y}_1 - \mathcal{Y}_0 \right ]
\end{equation}
\begin{equation}
    CATE = \mathbb{E}\left [ \mathcal{Y}_1|X=x \right ] - \mathbb{E}\left [ \mathcal{Y}_0|X=x \right ]
\end{equation}
and $\mathbb{E}[.]$ denotes mathematical expectation. The ATE is essentially the average ITE for the entire population; the CATE is the average ITE for everyone in the subpopulation characterised by $X=x$. The ATE is not meaningful if there is substantial heterogeneity of the ITEs between subpopulations. In such circumstances, CATE is more informative about ITEs as it allows the effect to be conditioned on the subpopulation of interest. The ITE can be thought of as a special case of CATE where individual $i$ is the only member of the subpopulation. While $ITE_i$ cannot be identified, $CATE$ for the subpopulation $X=x$ which includes individual $i$ will be better estimate of it than $ATE$ (under the reasonable assumption that between-subpopulation variation in ITEs is greater than that within subpopulations).  

Despite the fact that the aforementioned treatment effects usually cannot be calculated directly, successful methods have been developed so far that attempt to approximate those quantities. Perhaps the simplest and most naive approach is regression adjustment, where a regressor, or multiple ones per each treatment value, is used to estimate potential outcomes. More advanced methods often incorporate propensity scores, where the estimator takes into account the probability of treatment assignment per each individual. For instance, Inverse Propensity Weighting \cite{rosenbaumCentralRolePropensity1983} adjusts sample importances, further extended to more efficient and stable Doubly Robust method \cite{robinsEstimationRegressionCoefficients1994, fosterOrthogonalStatisticalLearning2020}. Double Machine Learning \cite{chernozhukovDoubleDebiasedMachine2018}, on the other hand, improves existing statistical estimators using base learners. Furthermore, recent surge in machine learning also delivered powerful procedures, often pushing state-of-the-art results \cite{johanssonLearningRepresentationsCounterfactual2016,shalitEstimatingIndividualTreatment2017,yaoRepresentationLearningTreatment2018,louizosCausalEffectInference2017}. In the realm of ensembles, there is Causal Forest \cite{atheyGeneralizedRandomForests2019} that specifically targets CATE estimation. Another interesting perspective on the problem is given through meta-learners \cite{kunzelMetalearnersEstimatingHeterogeneous2019, nieQuasiOracleEstimationHeterogeneous2020}, where out of the box estimators are used in various combinations and strategies to collectively approximate causal effects.

These are the most common methods that employ the usual assumptions, that is, \textit{SUTVA} and \textit{strong ignorability}, though there are many procedures that attempt to relax some of the assumptions as well. Here, we limit our discussion to this standard set of assumptions as it is relevant to this work. For a broader overview of available causal inference methods, as well as formal definitions of the assumptions, consult recent reviews on the topic \cite{guoSurveyLearningCausality2020, yaoSurveyCausalInference2020}.

\section{Model Misspecification}
The choice of model class occurs at some point in any learning task. Such a decision is made based on available data, usually the training part of it, while the environment of the actual application can be different, a scenario often mimicked via a separate test set. The occurring discrepancies between those two data sets are known as covariate shift problem. Within causal inference, this manifests as differences between observational and interventional distributions, ultimately making effect estimation extremely difficult. More formally, given input covariates $x$, treatment $t$, and outcome $y$, the conditional distribution $P(y|x, t)$ remains unchanged across the entire data set, whereas marginal distributions $P(x, t)$ differ between observational and interventional data. This is where model misspecification occurs as the model class is selected based on available observations only, which does not generalise well to later predicted interventions.

\begin{figure}[tb]
    \centering
    \includegraphics[width=0.46\textwidth]{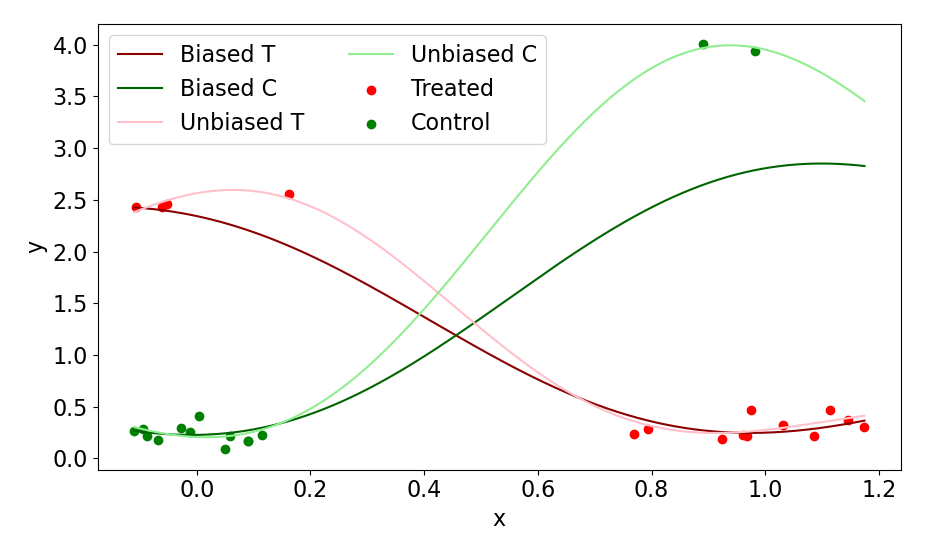}
    \caption{An example highlighting model misspecification issue. T and C denote Treated and Control respectively. The difference in ITE error is almost twice as in ATE.}
    \label{fig:misspec}
\end{figure}

Let us consider a simple example as presented in Figure \ref{fig:misspec}. It consists of a single input feature $x$, output variable $y$ (both continuous), and binary treatment $t$. For convenience, let us denote this data set as $\mathcal{D}$. Note the effect is clearly heterogeneous as it differs in $\mathcal{D}(x < 0.5)$ and $\mathcal{D}(x > 0.5)$. Furthermore, the two data regions closer to the top of the figure, that is, $\mathcal{D}(x < 0.5, t=1)$ and $\mathcal{D}(x > 0.5, t=0)$, are in minority with respect to the rest of the data. By many learners these scarce data points will likely be treated as outliers, resulting in lower variance than needed to provide accurate estimates. Thus, naively fitting the data will lead to biased estimates, an example of which is depicted on the figure as \textit{Biased T} and \textit{Biased C}. However, what we aim for is an unbiased estimator that captures the data closely while still generalising well, a scenario showcased by \textit{Unbiased T} and \textit{Unbiased C} on the figure.

For ITE estimation, fitting the data closely is especially important. Although in case of average effect estimation the difference between biased and unbiased estimators can be negligible, the individualised case usually exacerbates the issue. For instance, in the presented example, the difference in ATE error is $0.44$, but it grows to $0.77$ in ITE error.

In this work, instead of altering the sample importance, as many existing methods do, we aim to augment provided data in a way that underrepresented data regions are no longer dominated by the rest of the samples, leading to estimators no longer treating those data points as outliers and fitting them more closely, ultimately resulting in less biased solutions and more accurate ITE estimates. The following section describes our proposed method in detail.

\section{Debiasing Generative Trees}
As described in the previous section, model misspecification can be caused by underrepresented or missing data regions. Reweighing partially addresses this problem, but struggles with ITE estimation, not to mention propensity score approximators are subject to misspecification too. To avoid these pitfalls, we tackle the misspecification through undersmoothness by augmenting the original data with new data points that carry useful information and help achieve the final estimators better ITE predictions. As the injected samples are expected to be informative to the learners, the overall data complexity increases as a consequence. Moreover, because this is a data augmentation procedure, it is estimator agnostic, that is, it can be used by any existing estimation methods. It is also worth pointing out that simply modelling and oversampling the entire joint distribution would not work as the learnt joint would include any existing data imbalances. In other words, underrepresented data regions would remain in minority, not addressing the problem at hand.

This observation led us to a conclusion there is a need to identify smaller data regions, or clusters, and model their distributions in separation instead, giving us control over which areas to sample from and with what ratios. To achieve this, we incorporate recently proposed Generative Trees \cite{correiaJointsRandomForests2020a}, which retain all the benefits of standard decision trees, such as simplicity, speed and transparency. They can also be easily extended to ensembles of trees, often improving the performance significantly. In practice, a standard decision tree regressor is used to learn the data. Once the tree is constructed, the samples can be assigned to tree leaves according to the learnt decision paths, forming distinct subpopulations that we are after. The distributions of these clusters are then separately modelled through Gaussian Mixture Models (GMMs). Similarly to decision trees, we again prioritise simplicity and ease of use here, which is certainly the case with GMMs. The next step is to sample equally from modelled distributions, that is, to draw the same amount of new samples per each GMM. In this way, we reduce data imbalances. A merge of new and original data is then provided to a downstream estimator, resulting in a less biased final estimator. Through experimentation, we find that splitting the original data at the beginning of the process into treated and control units and learning two separate trees for each group helps achieve better overall effect. A step-by-step description of the proposed procedure is presented in Algorithm \ref{alg:dege}.

\begin{algorithm}[tb]
\caption{Debiasing Generative Trees}
\label{alg:dege}
\textbf{Input}: $X$ - data set, E - estimator\\
\textbf{Parameter}: N - number of generated samples\\
\textbf{Output}: $E_D$ - debiased estimator\\
\begin{algorithmic}[1] 
\STATE Let $X_G = \varnothing$.
\STATE Split $X$ into treated and control units ($X_T$ and $X_C$).
\STATE Train a Decision Tree regressor on $X_T$.
\STATE Map $X_T$ to tree leaves. Obtain subpopulations $S$.
\STATE Let $N_G = N/(2 \times len(S))$.
\FOR{$S_i$ in $S$}
\STATE Model $S_i$ with Gaussian Mixture Models. Obtain $G_i$.
\STATE Draw $N_G$ samples from $G_i$. Store them in $X_G$.
\ENDFOR
\STATE Repeat steps 3-9 for $X_C$.
\STATE Merge $X$ and $X_G$ into a single data set $X_M$.
\STATE Train estimator $E$ on $X_M$. Get debiased estimator $E_D$.
\STATE \textbf{return} debiased estimator $E_D$
\end{algorithmic}
\end{algorithm}

As ensembles of trees almost always improve over simple ones, we incorporate Extremely Randomised Trees for an additional performance gain. The procedure remains the same on a high level, differing only in randomly selecting inner trees at the time of sampling. Overall, we call this approach Debiasing Generative Trees (DeGeTs) as a general framework, with DeGe Decision Trees (DeGeDTs) and DeGe Forests (DeGeFs) for realisations with Decision Trees and Extremely Randomised Trees respectively.

There are a few important parameters to take care of when using the method. Firstly, depth of trees controls the granularity of identified subpopulations. Smaller clusters may translate to less accurate modelled distributions, whereas too shallow trees will bring the modelling closer to the entire joint that may result in not solving the problem of interest at all. The other tunable knob is the amount of new data samples to generate, where more data usually equates to a stronger effect, but also higher noise levels, which must be controlled to avoid destroying meaningful information in the original data. Finally, the number of components in GMMs is worth considering, where more complex distributions may require higher numbers of components.

All of the parameters can be found through cross-validation by using a downstream estimator's performance as a feedback signal as to which parameters work the best, which can also be tailored to a specific estimator of choice. The number of GMM components can be alternatively optimised through Bayesian Information Criterion (BIC) score. In order to make this method as general and easy to use as possible, we instead provide a set of reasonable defaults that we find work well across different data sets and settings. Default parameters: $max\_depth = \left \lceil \log_2 N_f \right \rceil - 1$, where $N_f$ denotes the number of input features, $n\_samples = 0.5 \times size(training\_data)$, $n\_components \in [1, 5]$ --- pick the one with the lowest BIC score.

In addition, we observe the fact that DeGeTs framework goes beyond applied Generative Trees and GMMs. This is because the data splitting part can, in fact, be performed by other methods, such as clustering. Consequently, GMMs can be substituted by any other generative models.

\section{Experiments}
We follow recent literature (e.g. \cite{johanssonLearningRepresentationsCounterfactual2016, shalitEstimatingIndividualTreatment2017, yaoRepresentationLearningTreatment2018}) in terms of incorporated data sets and evaluation metrics. We start with defining the later as different data sets use different sets of metrics. The source code that allows for a full replication of the presented experiments is available online\footnote{https://github.com/misoc-mml/undersmoothing-data-augmentation} and is based on the \textit{CATE benchmark}\footnote{https://github.com/misoc-mml/cate-benchmark}.

There are a few aspects we aim to investigate. Firstly, how the established reweighing methods perform in individual treatment effect estimation. Secondly, how the choice of model class impacts estimation accuracy (misspecification). Thirdly, how our proposed method affects the performance of the base learners, and how it compares to other methods. Finally, we also study how our method influences the number of rules in prunned decision trees as an indirect measure of data complexity.

Although we do perform hyperparameter search to some extent in order to get reasonable results, it is not our goal to achieve the best results possible, hence the parameters used here are likely not optimal and can be improved upon more extensive search. The main reason is the setups presented as part of this work are intended to be as general as possible. This is why in our analysis we specifically focus on the relative difference in performance between settings rather than comparing them to absolute state-of-the-art results.

\subsection{Evaluation Metrics}
The main focus of utilised metrics here is on the quantification of the errors made by provided predictions. Thus, the metrics are usually denoted as $\epsilon_X$, which translates to the amount of error made with respect to prediction type $X$ (lower is better). In terms of treatment outcomes, $\mathcal{Y}_t^{(i)}$ and $\hat{y}_t^{(i)}$ denote true and predicted outcomes respectively for treatment $t$ and individual $i$. Thus, following the definition of ITE (Eq. (\ref{eq:ite})), the difference $\mathcal{Y}_1^{(i)} - \mathcal{Y}_0^{(i)}$ gives a true effect, whereas $\hat{y}_1^{(i)} - \hat{y}_0^{(i)}$ a predicted one. Following this, we can define Precision in Estimation of Heterogeneous Effect (PEHE), which is the root mean squared error between predicted and true effects:
\begin{equation}\label{eq:e_pehe}
    \epsilon_{PEHE}=\sqrt{\frac{1}{n}\sum_{i=1}^{n}( \hat{y}_1^{(i)}  - \hat{y}_0^{(i)} - (\mathcal{Y}_1^{(i)} - \mathcal{Y}_0^{(i)}))^2}
\end{equation}

Following the definition of ATE (Eq. (\ref{eq:ate})), we measure the error on ATE as the absolute difference between predicted and true average effects, formally written as:
\begin{equation}\label{eq:e_ate}
    \epsilon_{ATE}=\left | \frac{1}{n} \sum_{i=1}^{n} (\hat{y}_1^{(i)} - \hat{y}_0^{(i)}) - \frac{1}{n} \sum_{i=1}^{n} (\mathcal{Y}_1^{(i)} - \mathcal{Y}_0^{(i)}) \right |
\end{equation}

Given a set of treated subjects $T$ that are part of sample $E$ coming from an experimental study, and a set of control group $C$, define the true Average Treatment effect on the Treated (ATT) as:
\begin{equation}
    ATT = \frac{1}{|T|}\sum_{i \in T} \mathcal{Y}^{(i)} - \frac{1}{|C \cap E|}\sum_{i \in C \cap E} \mathcal{Y}^{(i)}
\end{equation}

The error on ATT is then defined as the absolute difference between the true and predicted ATT:
\begin{equation}\label{eq:e_att}
    \epsilon_{ATT} = \left | ATT - \frac{1}{|T|} \sum_{i \in T} (\hat{y}_1^{(i)} - \hat{y}_0^{(i)}) \right |
\end{equation}

Define policy risk as:
\begin{multline}\label{eq:r_pol}
    \mathcal{R}_{pol} = 1 - (\mathbb{E}\left [ \mathcal{Y}_1|\pi(x)=1 \right ] \mathcal{P}(\pi(x)=1) \\ 
    + \mathbb{E}\left [ \mathcal{Y}_0|\pi(x)=0 \right ] \mathcal{P}(\pi(x)=0)
\end{multline}
Where $\mathbb{E}[.]$ denotes mathematical expectation and policy $\pi$ becomes $\pi(x)=1$ if $\hat{y}_1 - \hat{y}_0 > 0$; $\pi(x)=0$ otherwise.

\subsection{Data}
We incorporate a set of well-established causal inference benchmark data sets that are briefly described in the following paragraphs and summarised in Table \ref{tab:data}.

\textbf{IHDP}. Introduced by \cite{hillBayesianNonparametricModeling2011}, based on Infant Health Development Program (IHDP) clinical trial \cite{brooks-gunnEffectsEarlyIntervention1992}. The experiment measured various aspects of premature infants and their mothers, and how receiving specialised childcare affected the cognitive test score of the infants later on. We use a semi-synthetic version of this data set, where the outcomes are simulated through the NPCI package\footnote{https://github.com/vdorie/npci} (setting `A') based on real pre-treatment covariates. Moreover, the treatment groups are made imbalanced by removing a subset of the treated individuals. We report errors on estimated PEHE and ATE averaged over 1,000 realisations and split the data with 90/10 training/test ratios.

\textbf{JOBS}. This data set, proposed by \cite{a.smithDoesMatchingOvercome2005}, is a combination of the experiment done by \cite{lalondeEvaluatingEconometricEvaluations1986} as part of the National Supported Work Program (NSWP) and observational data from the Panel Study of Income Dynamics (PSID) \cite{dehejiaPropensityScoreMatchingMethods2002}. Overall, the data captures people's basic characteristics, whether they received a job training from NSWP (treatment), and their employment status (outcome). Here, we report $\epsilon_{ATT}$ and $\mathcal{R}_{pol}$ averaged over 10 runs with 80/20 training/test ratio splits.

\textbf{NEWS}. Introduced by \cite{johanssonLearningRepresentationsCounterfactual2016}, which consists of news articles in the form of word counts with respect to a predefined vocabulary. The treatment is represented as the device type (mobile or desktop) used to view the article, whereas the simulated outcome is defined as the user's experience. Similarly to IHDP, we report PEHE and ATE errors for this data set, averaging over 50 realisations with 90/10 training/test ratio splits.

\textbf{TWINS}. The data set comes from official records of twin births in the US in years 1989-1991 \cite{almondCostsLowBirth2005}. The data are preprocessed to include only individuals of the same sex and where each of them weight less than 2,000 grams. The treatment is represented as whether the individual is the heavier one of the twins, whereas the outcome is the mortality within the first year of life. As both factual and counterfactual outcomes are known from the official records, that is, mortality of both twins, one of the twins is intentionally hidden to simulate an observational setting. Here, we incorporate the approach taken by \cite{louizosCausalEffectInference2017}, where new binary features are created and flipped at random ($0.33$ probability) in order to hide confounding information. We report $\epsilon_{ATE}$ and $\epsilon_{PEHE}$ for this data set, averaged over 10 iterations with 80/20 training/test ratio splits.

\begin{table}[t]
    \centering
    \caption{A summary of incorporated data sets. t/c denote the amount of treated and control samples respectively.}
    \begin{tabular}{llll}
    \toprule
        data set & \# samples (t/c) & \# features & outcome \\
    \midrule
        IHDP & 747 (139/608) & 25 & cont. \\
        JOBS & 3,212 (297/2,915) & 17 & binary \\
        NEWS & 5,000 (2,289/2,711) & 3,477 & cont. \\
        TWINS & 11,984 (5,992/5,992) & 194 & binary \\
    \bottomrule
    \end{tabular}
    \label{tab:data}
\end{table}

\subsection{Setup}
We incorporate the following estimators.

\textbf{Base Learners}. Linear methods: Lasso (l1) and Ridge (l2). Simple Trees: prunned Decision Trees, Extremely Randomised Trees (ET) \cite{geurtsExtremelyRandomizedTrees2006}. Gradient Boosted Trees: CatBoost\footnote{https://github.com/catboost/catboost}, LightGBM \cite{keLightGBMHighlyEfficient2017}. Kernel Ridge regression with nonlinearities. Dummy regressor returning the mean as a reference only.

\textbf{Reweighing Methods}. Causal Forest \cite{atheyGeneralizedRandomForests2019}, Double Machine Learning (DML) \cite{chernozhukovDoubleDebiasedMachine2018}, and Meta-Learners \cite{kunzelMetalearnersEstimatingHeterogeneous2019} in the form of T and X variations.

\textbf{Debiasing Generative Trees}. Our proposed method. We include the stronger performing DeGeF variation.

A general approach throughout all conducted experiments was to train a method on the training set and evaluate it against appropriate metrics on the test set. 5 base learners were trained and evaluated in that way: l1, l2, Simple Trees, Boosted Trees and Kernel Ridge. DML and Meta-Learners were combined with different base learners as they need them to solve intermediate regression and classification tasks internally. This resulted in $3 \times 5 = 15$ combinations of distinct estimators. Similarly, DeGeF was combined with the same 5 base learners to investigate how they react to our data augmentation method. Causal Forest and dummy regressor were treated as standalone methods. Overall, we obtained 27 distinct estimators per each data set. In terms of Simple and Boosted Trees, we defaulted to ETs and CatBoost respectively. For NEWS, due to its high-dimensionality, we switched to computationally less expensive Decision Trees and LightGBM instead.

As our DeGeF method is a data augmentation approach, it affects only the training set that is later used by base learners. It does not change the test set in any way as the test portion is used specifically for evaluation purposes to test how methods generalise to unseen data examples. More specifically, DeGeF injects new data samples to the existing training set, and that augmented training set is then provided to base learners.

Hyperparameter search was also performed wherever applicable, though not too extensive to keep our study as general and accessible as possible. The following is a list of base learners and their hyperparameters we explored. ETs: \textit{max\_leaf\_nodes} $\in \{10, 20, 30, None\}$, \textit{max\_depth} $\in \{5, 10, 20\}$. Kernel Ridge: \textit{alpha} $\in \{0, 1e-1, 1e-2, 1e-3 \}$, \textit{gamma} $\in \{1e-2, 1e-1, 0, 1e+1, 1e+2 \}$, \textit{kernel} $\in \{rbf, poly \}$, \textit{degree} $\in \{2, 3, 4 \}$. CatBoost: \textit{depth} $\in \{6, 8, 10 \}$, \textit{l2\_leaf\_reg} $\in \{1, 3, 10, 100 \}$. LightGBM: \textit{max\_depth} $\in \{5, 7, 10 \}$, \textit{reg\_lambda} $\in \{0, 0.1, 1, 5, 10 \}$. Causal Forest: \textit{max\_depth} $\in \{5, 10, 20 \}$. For ETs, CatBoost, LightGBM and Causal Forest we set the number of inner estimators to $1000$. To find the best set of hyperparameters, we performed 5-fold cross-validation. When it comes to DeGeF, we set the number of estimators to $10$. The other parameters, like number of new samples, tree depth and GMM components, were set to defaults as recommended in the description of the framework. All randomisation seeds were set to a fixed number ($1$) throughout all experiments.

Most of our experimental runs were performed on a Linux based machine with 12 CPUs and 60 GBs of RAM. More demanding settings, such as NEWS combined with tree-based methods, were delegated to one with 96 CPUs and 500 GBs of RAM, though such a powerful machine is not required to complete those runs.

\subsection{Results}
We incorporate the following estimator names throughout the presented tables: \textbf{l1} - Lasso, \textbf{l2} - Ridge, \textbf{kr} - Kernel Ridge, \textbf{dt} - Decision Tree, \textbf{et} - Extremely Randomised Trees, \textbf{cb} - CatBoost, \textbf{lgbm} - LightGBM, \textbf{cf} - Causal Forest, \textbf{dml} - Double Machine Learning, \textbf{xl} - X-Learner, \textbf{degef} - our DeGeF method. Combinations of the methods are denoted with a hyphen, for instance, `dml-l1'.

Tables \ref{tab:ihdp} - \ref{tab:news} present the main results, where we specifically focus on: a) relevant to a given data set metrics, and b) changes in performance relative to a particular base learner. The latter is calculated as $((r_a - r_b) / r_b) \times 100\%$, where $r_a$ and $r_b$ denote results of advanced methods and base learners respectively. The reason for analysing these relative changes rather than absolute values is because in this study we are specifically interested in how more complex approaches (including ours) affect the performance of the base learners, even if not reaching state-of-the-art results. For example, if a relative change for \textit{xl-et} reads `$-20$', it means this estimator decreased the error by $20\%$ when compared to plain \textit{et} learner for that particular metric. Changes greater than zero denote an increase in errors (lower is better). Furthermore, Table \ref{tab:rules_count} shows the number of rules obtained from a prunned Decision Tree while trained on original data and augmented by \textit{degef}. All presented numbers (excluding relative percentages) denote means and $95\%$ confidence intervals.

\begin{table}[tb]
    \centering
    \caption{IHDP results.}
    \begin{tabular}{llrlr}
\toprule
      name &         $\epsilon_{ATE}$ & $\Delta\%$ &        $\epsilon_{PEHE}$ & $\Delta\%$ \\
\midrule
     dummy &  $4.408\pm0.103$ &               - &  $7.898\pm0.473$ &                - \\
     l1 &  $0.981\pm0.106$ &               - &  $5.790\pm0.514$ &                - \\
     l2 &  $0.974\pm0.104$ &               - &  $5.786\pm0.514$ &                - \\
        dt &  $0.636\pm0.084$ &               - &  $4.025\pm0.402$ &                - \\
        kr &  $0.356\pm0.031$ &               - &  $2.276\pm0.170$ &                - \\
        et &  $0.519\pm0.074$ &               - &  $3.093\pm0.322$ &                - \\
        cb &  $0.404\pm0.038$ &               - &  $2.179\pm0.210$ &                - \\
      lgbm &  $0.412\pm0.052$ &               - &  $2.866\pm0.273$ &                - \\
        cf &  $0.397\pm0.045$ &               - &  $3.387\pm0.318$ &                - \\
 dml-l1 &  $0.387\pm0.043$ &      $-60.52$ &  $7.782\pm0.691$ &        $34.42$ \\
 dml-l2 &  $0.381\pm0.040$ &      $-60.91$ &  $7.859\pm0.691$ &        $35.82$ \\
    dml-dt &  $1.262\pm0.116$ &       $98.50$ &  $6.679\pm0.570$ &        $65.95$ \\
    dml-kr &  $0.616\pm0.059$ &       $73.06$ &  $8.174\pm0.728$ &       $259.16$ \\
    dml-et &  $0.869\pm0.082$ &       $67.61$ &  $6.532\pm0.563$ &       $111.23$ \\
    dml-cb &  $1.123\pm0.052$ &      $177.88$ &  $6.976\pm0.580$ &       $220.18$ \\
  dml-lgbm &  $1.516\pm0.142$ &      $268.30$ &  $7.544\pm0.632$ &       $163.25$ \\
  tl-l1 &  $0.273\pm0.033$ &      $-72.19$ &  $7.858\pm0.678$ &        $35.73$ \\
  tl-l2 &  $0.273\pm0.034$ &      $-72.02$ &  $7.810\pm0.679$ &        $34.99$ \\
     tl-dt &  $0.406\pm0.044$ &      $-36.22$ &  $8.012\pm0.698$ &        $99.07$ \\
     tl-kr &  $0.167\pm0.010$ &      $-53.02$ &  $8.024\pm0.706$ &       $252.60$ \\
     tl-et &  $0.306\pm0.042$ &      $-41.01$ &  $7.445\pm0.643$ &       $140.75$ \\
     tl-cb &  $0.224\pm0.027$ &      $-44.48$ &  $7.715\pm0.664$ &       $254.10$ \\
   tl-lgbm &  $0.255\pm0.028$ &      $-38.10$ &  $8.002\pm0.678$ &       $179.25$ \\
  xl-l1 &  $0.282\pm0.034$ &      $-71.27$ &  $7.660\pm0.678$ &        $32.31$ \\
  xl-l2 &  $0.287\pm0.034$ &      $-70.53$ &  $7.723\pm0.678$ &        $33.47$ \\
     xl-dt &  $0.529\pm0.065$ &      $-16.81$ &  $7.317\pm0.653$ &        $81.79$ \\
     xl-kr &  $0.247\pm0.023$ &      $-30.65$ &  $7.847\pm0.698$ &       $244.82$ \\
     xl-et &  $0.453\pm0.053$ &      $-12.63$ &  $6.875\pm0.597$ &       $122.32$ \\
     xl-cb &  $0.388\pm0.044$ &       $-3.97$ &  $6.894\pm0.604$ &       $216.42$ \\
   xl-lgbm &  $0.435\pm0.046$ &        $5.53$ &  $7.602\pm0.650$ &       $165.29$ \\
 degef-l1 &  $1.051\pm0.107$ &        $7.15$ &  $5.809\pm0.514$ &         $0.33$ \\
 degef-l2 &  $1.093\pm0.107$ &       $12.16$ &  $5.820\pm0.514$ &         $0.58$ \\
    degef-dt &  $0.542\pm0.075$ &      $-14.83$ &  $3.882\pm0.384$ &        $-3.55$ \\
    degef-kr &  $0.316\pm0.031$ &      $-11.18$ &  $2.149\pm0.181$ &        $-5.58$ \\
    degef-et &  $0.394\pm0.052$ &      $-24.03$ &  $2.818\pm0.273$ &        $-8.89$ \\
    degef-cb &  $0.328\pm0.032$ &      $-18.73$ &  $2.013\pm0.190$ &        $-7.63$ \\
  degef-lgbm &  $0.397\pm0.051$ &       $-3.54$ &  $2.691\pm0.250$ &        $-6.09$ \\
\bottomrule
\end{tabular}
    \label{tab:ihdp}
\end{table}

\begin{table}[tb]
    \centering
    \caption{JOBS results.}
    \begin{tabular}{llrlr}
\toprule
      name &         $\epsilon_{ATT}$ & $\Delta\%$ &      $\mathcal{R}_{pol}$ & $\Delta\%$ \\
\midrule
     dummy &  $0.029\pm0.000$ &               - &  $0.326\pm0.000$ &                  - \\
     l1 &  $0.005\pm0.000$ &               - &  $0.296\pm0.000$ &                  - \\
     l2 &  $0.034\pm0.000$ &               - &  $0.296\pm0.000$ &                  - \\
        dt &  $0.029\pm0.000$ &               - &  $0.365\pm0.000$ &                  - \\
        kr &  $0.017\pm0.000$ &               - &  $0.400\pm0.000$ &                  - \\
        et &  $0.006\pm0.000$ &               - &  $0.276\pm0.000$ &                  - \\
        cb &  $0.026\pm0.000$ &               - &  $0.308\pm0.000$ &                  - \\
      lgbm &  $0.029\pm0.000$ &               - &  $0.247\pm0.000$ &                  - \\
        cf &  $0.025\pm0.000$ &               - &  $0.294\pm0.000$ &                  - \\
 dml-l1 &  $0.012\pm0.000$ &      $146.75$ &  $0.366\pm0.000$ &          $23.43$ \\
 dml-l2 &  $0.008\pm0.000$ &      $-77.14$ &  $0.374\pm0.000$ &          $26.20$ \\
    dml-dt &  $0.149\pm0.000$ &      $408.57$ &  $0.336\pm0.000$ &          $-7.80$ \\
    dml-kr &  $0.007\pm0.000$ &      $-61.39$ &  $0.374\pm0.000$ &          $-6.52$ \\
    dml-et &  $0.099\pm0.000$ &     $1686.11$ &  $0.353\pm0.000$ &          $27.66$ \\
    dml-cb &  $0.010\pm0.000$ &      $-60.23$ &  $0.368\pm0.000$ &          $19.42$ \\
  dml-lgbm &  $0.191\pm0.000$ &      $555.20$ &  $0.387\pm0.000$ &          $56.81$ \\
  tl-l1 &  $0.012\pm0.000$ &      $140.15$ &  $0.374\pm0.000$ &          $26.10$ \\
  tl-l2 &  $0.007\pm0.000$ &      $-79.25$ &  $0.370\pm0.000$ &          $24.75$ \\
     tl-dt &  $0.035\pm0.000$ &       $21.07$ &  $0.351\pm0.000$ &          $-3.68$ \\
     tl-kr &  $0.005\pm0.000$ &      $-70.54$ &  $0.305\pm0.000$ &         $-23.81$ \\
     tl-et &  $0.010\pm0.000$ &       $86.50$ &  $0.295\pm0.000$ &           $6.81$ \\
     tl-cb &  $0.026\pm0.000$ &       $-0.50$ &  $0.250\pm0.000$ &         $-18.86$ \\
   tl-lgbm &  $0.004\pm0.000$ &      $-86.33$ &  $0.305\pm0.000$ &          $23.62$ \\
  xl-l1 &  $0.022\pm0.000$ &      $361.49$ &  $0.356\pm0.000$ &          $20.16$ \\
  xl-l2 &  $0.011\pm0.000$ &      $-67.37$ &  $0.361\pm0.000$ &          $21.91$ \\
     xl-dt &  $0.037\pm0.000$ &       $27.98$ &  $0.296\pm0.000$ &         $-18.75$ \\
     xl-kr &  $0.003\pm0.000$ &      $-80.58$ &  $0.279\pm0.000$ &         $-30.32$ \\
     xl-et &  $0.004\pm0.000$ &      $-36.17$ &  $0.235\pm0.000$ &         $-14.87$ \\
     xl-cb &  $0.045\pm0.000$ &       $72.93$ &  $0.239\pm0.000$ &         $-22.56$ \\
   xl-lgbm &  $0.021\pm0.000$ &      $-29.31$ &  $0.297\pm0.000$ &          $20.20$ \\
 degef-l1 &  $0.054\pm0.012$ &     $1010.26$ &  $0.296\pm0.000$ &           $0.00$ \\
 degef-l2 &  $0.056\pm0.009$ &       $62.76$ &  $0.296\pm0.000$ &           $0.00$ \\
    degef-dt &  $0.048\pm0.014$ &       $64.01$ &  $0.335\pm0.015$ &          $-8.12$ \\
    degef-kr &  $0.019\pm0.012$ &       $11.82$ &  $0.299\pm0.013$ &         $-25.16$ \\
    degef-et &  $0.015\pm0.009$ &      $167.98$ &  $0.270\pm0.014$ &          $-2.24$ \\
    degef-cb &  $0.019\pm0.007$ &      $-26.61$ &  $0.257\pm0.030$ &         $-16.51$ \\
  degef-lgbm &  $0.021\pm0.007$ &      $-27.62$ &  $0.283\pm0.024$ &          $14.83$ \\
\bottomrule
\end{tabular}
    \label{tab:jobs}
\end{table}

\begin{table}[tb]
    \centering
    \caption{TWINS results.}
    \begin{tabular}{llrlr}
\toprule
      name &         $\epsilon_{ATE}$ & $\Delta\%$ &        $\epsilon_{PEHE}$ & $\Delta\%$ \\
\midrule
     dummy &  $0.033\pm0.002$ &               - &  $0.318\pm0.004$ &                - \\
     l1 &  $0.042\pm0.000$ &               - &  $0.319\pm0.004$ &                - \\
     l2 &  $0.047\pm0.002$ &               - &  $0.320\pm0.004$ &                - \\
        dt &  $0.004\pm0.005$ &               - &  $0.319\pm0.004$ &                - \\
        kr &  $0.045\pm0.001$ &               - &  $0.320\pm0.004$ &                - \\
        et &  $0.027\pm0.006$ &               - &  $0.322\pm0.003$ &                - \\
        cb &  $0.039\pm0.000$ &               - &  $0.319\pm0.004$ &                - \\
      lgbm &  $0.038\pm0.000$ &               - &  $0.327\pm0.005$ &                - \\
        cf &  $0.064\pm0.001$ &               - &  $0.323\pm0.005$ &                - \\
 dml-l1 &  $0.028\pm0.003$ &      $-33.55$ &  $0.318\pm0.004$ &        $-0.29$ \\
 dml-l2 &  $0.042\pm0.001$ &      $-11.32$ &  $0.334\pm0.009$ &         $4.25$ \\
    dml-dt &  $0.070\pm0.011$ &     $1859.14$ &  $0.327\pm0.002$ &         $2.53$ \\
    dml-kr &  $0.055\pm0.028$ &       $20.87$ &  $0.323\pm0.012$ &         $0.99$ \\
    dml-et &  $0.047\pm0.002$ &       $74.36$ &  $0.320\pm0.005$ &        $-0.32$ \\
    dml-cb &  $0.078\pm0.011$ &       $99.66$ &  $0.328\pm0.002$ &         $2.65$ \\
  dml-lgbm &  $0.034\pm0.007$ &      $-10.56$ &  $0.362\pm0.008$ &        $10.90$ \\
  tl-l1 &  $0.052\pm0.001$ &       $23.80$ &  $0.324\pm0.005$ &         $1.59$ \\
  tl-l2 &  $0.042\pm0.000$ &      $-10.47$ &  $0.337\pm0.011$ &         $5.19$ \\
     tl-dt &  $0.062\pm0.000$ &     $1631.81$ &  $0.334\pm0.004$ &         $4.67$ \\
     tl-kr &  $0.050\pm0.000$ &        $9.18$ &  $0.334\pm0.006$ &         $4.45$ \\
     tl-et &  $0.051\pm0.000$ &       $87.25$ &  $0.327\pm0.006$ &         $1.76$ \\
     tl-cb &  $0.051\pm0.000$ &       $31.77$ &  $0.331\pm0.008$ &         $3.65$ \\
   tl-lgbm &  $0.042\pm0.002$ &        $9.79$ &  $0.393\pm0.009$ &        $20.34$ \\
  xl-l1 &  $0.053\pm0.001$ &       $25.46$ &  $0.322\pm0.004$ &         $0.71$ \\
  xl-l2 &  $0.042\pm0.001$ &      $-10.95$ &  $0.335\pm0.010$ &         $4.83$ \\
     xl-dt &  $0.059\pm0.000$ &     $1549.54$ &  $0.323\pm0.004$ &         $1.20$ \\
     xl-kr &  $0.043\pm0.002$ &       $-4.96$ &  $0.325\pm0.007$ &         $1.73$ \\
     xl-et &  $0.050\pm0.001$ &       $85.14$ &  $0.323\pm0.006$ &         $0.53$ \\
     xl-cb &  $0.048\pm0.002$ &       $22.63$ &  $0.323\pm0.006$ &         $1.04$ \\
   xl-lgbm &  $0.039\pm0.002$ &        $2.02$ &  $0.366\pm0.009$ &        $12.18$ \\
 degef-l1 &  $0.064\pm0.004$ &       $53.10$ &  $0.323\pm0.004$ &         $1.18$ \\
 degef-l2 &  $0.067\pm0.004$ &       $41.28$ &  $0.324\pm0.004$ &         $1.10$ \\
    degef-dt &  $0.064\pm0.013$ &     $1697.62$ &  $0.349\pm0.005$ &         $9.37$ \\
    degef-kr &  $0.033\pm0.004$ &      $-27.17$ &  $0.320\pm0.004$ &         $0.15$ \\
    degef-et &  $0.054\pm0.007$ &       $96.91$ &  $0.335\pm0.002$ &         $4.23$ \\
    degef-cb &  $0.051\pm0.003$ &       $31.48$ &  $0.326\pm0.004$ &         $2.06$ \\
  degef-lgbm &  $0.042\pm0.002$ &        $8.61$ &  $0.328\pm0.006$ &         $0.56$ \\
\bottomrule
\end{tabular}
    \label{tab:twins}
\end{table}

\begin{table}[tb]
    \centering
    \caption{NEWS results. Estimators marked with `x' -- no results due to unreasonably excessive training time.}
    \begin{tabular}{llrlr}
\toprule
      name &         $\epsilon_{ATE}$ & $\Delta\%$ &        $\epsilon_{PEHE}$ & $\Delta\%$ \\
\midrule
     dummy &  $2.714\pm0.212$ &               - &  $4.381\pm0.361$ &                - \\
     l1 &  $0.244\pm0.068$ &               - &  $3.370\pm0.365$ &                - \\
     l2 &  $0.260\pm0.068$ &               - &  $3.371\pm0.366$ &                - \\
        dt &  $0.344\pm0.076$ &               - &  $2.717\pm0.277$ &                - \\
        kr &  $0.715\pm0.133$ &               - &  $3.316\pm0.367$ &                - \\
        et &  $0.276\pm0.051$ &               - &  $2.063\pm0.200$ &                - \\
        cb &  $0.127\pm0.029$ &               - &  $1.880\pm0.179$ &                - \\
      lgbm &  $0.162\pm0.045$ &               - &  $2.074\pm0.241$ &                - \\
        cf &  $0.544\pm0.089$ &               - &  $3.907\pm0.481$ &                - \\
 dml-l1 &  $0.233\pm0.062$ &       $-4.50$ &  $2.469\pm0.269$ &       $-26.73$ \\
 dml-l2 &  $0.236\pm0.080$ &       $-9.08$ &  $5.108\pm0.394$ &        $51.52$ \\
    dml-dt &  $4.523\pm0.783$ &     $1216.23$ &  $5.875\pm0.676$ &       $116.18$ \\
    dml-kr &  $2.544\pm0.256$ &      $255.79$ &  $4.186\pm0.399$ &        $26.25$ \\
    dml-et &  x &      - &  x &        - \\
    dml-cb &  x &      - &  x &        - \\
  dml-lgbm &  $1.461\pm0.181$ &      $799.12$ &  $3.240\pm0.386$ &        $56.27$ \\
  tl-l1 &  $0.298\pm0.052$ &       $22.13$ &  $2.166\pm0.201$ &       $-35.74$ \\
  tl-l2 &  $0.173\pm0.030$ &      $-33.33$ &  $4.182\pm0.343$ &        $24.06$ \\
     tl-dt &  $0.329\pm0.062$ &       $-4.12$ &  $2.638\pm0.222$ &        $-2.92$ \\
     tl-kr &  $0.198\pm0.150$ &      $-72.27$ &  $2.677\pm0.290$ &       $-19.26$ \\
     tl-et &  x &      - &  x &        - \\
     tl-cb &  x &      - &  x &        - \\
   tl-lgbm &  $0.161\pm0.033$ &       $-0.81$ &  $1.861\pm0.138$ &       $-10.25$ \\
  xl-l1 &  $0.220\pm0.045$ &       $-9.75$ &  $2.152\pm0.186$ &       $-36.14$ \\
  xl-l2 &  $0.174\pm0.036$ &      $-33.09$ &  $4.162\pm0.345$ &        $23.45$ \\
     xl-dt &  $0.290\pm0.060$ &      $-15.47$ &  $2.639\pm0.263$ &        $-2.87$ \\
     xl-kr &  $0.229\pm0.112$ &      $-68.00$ &  $2.695\pm0.297$ &       $-18.72$ \\
     xl-et &  x &      - &  x &        - \\
     xl-cb &  x &      - &  x &        - \\
   xl-lgbm &  $0.131\pm0.042$ &      $-19.41$ &  $2.005\pm0.228$ &        $-3.31$ \\
 degef-l1 &  $0.225\pm0.048$ &       $-7.86$ &  $3.370\pm0.361$ &         $0.00$ \\
 degef-l2 &  $0.178\pm0.041$ &      $-31.64$ &  $3.366\pm0.362$ &        $-0.16$ \\
    degef-dt &  $0.355\pm0.080$ &        $3.22$ &  $2.727\pm0.266$ &         $0.35$ \\
    degef-kr &  $0.582\pm0.102$ &      $-18.61$ &  $3.256\pm0.349$ &        $-1.80$ \\
    degef-et &  $0.290\pm0.052$ &        $5.13$ &  $2.013\pm0.167$ &        $-2.40$ \\
    degef-cb &  x &      - &  x &        - \\
  degef-lgbm &  $0.151\pm0.042$ &       $-6.78$ &  $2.038\pm0.228$ &        $-1.71$ \\
\bottomrule
\end{tabular}
    \label{tab:news}
\end{table}

\section{Discussion}
In terms of IHDP data set (Table \ref{tab:ihdp}), the classic methods (\textit{dml}, \textit{tl}, and \textit{xl}) strongly improve in ATE, but can also be unstable as it is the case with \textit{dml}, specifically \textit{dml-cb} and \textit{dml-lgbm}. Against PEHE, the situation is much worse as those methods significantly decrease in performance when compared to the base learners, not to mention catastrophic setbacks in the worst cases (deltas above $200\%$). Note that not a single traditional method improves in PEHE (all deltas positive). Our \textit{degef}, on the other hand, often improves in both ATE and PEHE (see negative deltas). Even in the worst cases with \textit{l1} and \textit{l2}, \textit{degef} is still very stable and does not destroy the predictions as it happened with the other approaches. Thus, our method clearly offers the best improvements in PEHE and competitive predictions in ATE while providing a good amount of stability.

In the JOBS data set (Table \ref{tab:jobs}), classic methods again achieve strong improvements in average effect estimation (ATT) in best cases, though they can be substantially worse as well (e.g. \textit{dml-et}). In policy predictions, an equivalent of ITE, traditional techniques are even less likely to provide improvements, except the \textit{X-Learner}. With respect to \textit{degef}, it can also worsen the quality of predictions in ATT, as shown with \textit{degef-l1}, though it does not get as bad as with \textit{dml-et}. However, even in that worst example, policy predictions are not destroyed. The best cases in \textit{degef}, on the other hand, achieve strong improvements in policy. Similarly to IHDP, here \textit{degef} provided solid improvements in ITE predictions (policy), while staying on par with traditional methods in ATT, obtaining reasonable improvements and keeping the worst cases still better than the worst ones in the other methods, proving again its stability.

TWINS data set (Table \ref{tab:twins}), proved to be very difficult for all considered methods when it comes to PEHE, though they did not worsen the predictions as well. Some good improvements in ATE can be observed, but also noticeable decreases in performance in the worst cases (combinations with \textit{dt}). Our method behaves similarly to the classic ones, offering occasional gains and keeping the decreases in reasonable bounds. The stability of \textit{degef} is especially noticeable in PEHE as the worst decrease (\textit{degef-dt}) is still better than in other methods.

The last data set, NEWS (Table \ref{tab:news}), showed the traditional approaches can provide some improvements in PEHE as well, at least in their best efforts, though performance decreases are also noticeable in the worst ones. They also offer quite stable improvements in ATE, except extremely poor \textit{dml-dt}. The \textit{X-Learner} performs particularly well across both metrics (most deltas negative). Our proposed method offers reasonable gains in ATE as well, while keeping performance decreases at bay even in the worst efforts. Even though \textit{degef} provides little improvement in PEHE, it does not destroy individualised predictions either. Overall, this data set showcases superior stability properties of \textit{degef} particularly well, making it a preferable choice if small but safe performance gains are desirable over potentially higher but riskier improvements.

In general terms, the results show that performance can vary substantially depending on the model class, even within the same advanced method (\textit{dml, xl, degef}). For instance, \textit{DML} proved to work particularly well with \textit{L1} and \textit{L2} as base learners, whereas \textit{X-Learner} often outperforms \textit{T-Learner}, adding more stability to the results as well. Our proposed technique usually offers significant improvements in ITE predictions in best cases, often better than traditional methods, while keeping the predictions stable even in the worst examples. Classic methods are clearly strong in ATE estimates, but can struggle in individualised predictions. Overall, these methods (\textit{dml, xl}) proved to be less stable than ours, where the worst cases can perform quite poorly, especially \textit{dml}. This makes \textit{degef} a safer choice on average when considering various estimators, even more so when achieving the best possible performance is not considered a priority.

We also investigate the number of rules in prunned Decision Trees as a proxy for data complexity. As presented in Table \ref{tab:rules_count}, \textit{degef} significantly increases the amount of rules across all data sets, translating to an increase in data complexity. This proves the undersmoothing effect we aim for has been achieved. In addition, we observe that modest data complexity increases in IHDP and JOBS correlate with strong \textit{degef} gains in ITE estimation in those two data sets, whereas a much bigger difference in TWINS (from $9.6$ to $59.1$) correlated with considerably lower prediction performance gains (Table \ref{tab:twins}).

After combining all the results together, we can observe that \textit{degef}: a) improves effect predictions (Tables \ref{tab:ihdp} - \ref{tab:news}), and b) increases data complexity (Table \ref{tab:rules_count}). We thus conclude that more accurate effect prediction (as per a)) is a sign of better model generalisation. Consequently, we equate better generalisation to reduced model misspecification. Furthermore, we can observe the undersmoothing effect as per b). This is our indirect evidence that our method addresses misspecification via undersmoothing by showing that downstream estimators improve effect predictions when trained on augmented data. In terms of theoretical guarantees, we rely on \cite{whiteConsequencesDetectionMisspecified1981}, which provides a thorough formal analysis of the problem of undersmoothing.

In terms of possible limitations of our method, we assume the data sets we work with have relatively low noise levels. This is because in noisy environments, the inner GMMs would likely pick up a lot of noise and thus sampling from them would result in even more noisy data samples. The result would be the opposite of what we aim for, that is, to increase data complexity and bring new informative samples, not to introduce bias in the form of noise. Thus, our method would likely worsen base learners performance in such environments. Furthermore, we expect extremely high-dimensional data sets may cause computational issues due to the increasing depth of the inner trees. This is partly why setting a reasonable depth limit is important.

\begin{table}[t]
    \centering
    \caption{Number of rules in a prunned Decision Tree with and without \textit{degef} augmentation.}
    \begin{tabular}{lll}
\toprule
data set &            dt &         degef-dt \\
\midrule
    IHDP &  $33.6\pm2.0$ &   $53.3\pm2.6$ \\
    JOBS &   $6.0\pm0.0$ &   $11.3\pm5.3$ \\
   TWINS &   $9.6\pm0.9$ &  $59.1\pm11.9$ \\
    NEWS &  $19.4\pm2.5$ &   $32.0\pm4.7$ \\
\bottomrule
\end{tabular}
    \label{tab:rules_count}
\end{table}

\section{Conclusions}
Treatment effect estimation tasks are often subject to the covariate shift problem that is exhibited by discrepancies between observational and interventional distributions. This leads to model misspecification, which we tackle directly in this work by introducing a novel data augmentation method based on generative trees that provides an undersmoothing effect and helps downstream estimators achieve better robustness, ultimately leading to less biased estimators. Through our experiments, we show that the choice of model class matters, and that traditional methods can struggle in individualised effect estimation. Our proposed approach presented competitive results with existing reweighing procedures on average effect tasks while offering significantly better performance improvements on individual effect problems. The method also exhibits better stability in terms of provided gains than other approaches, rendering it a safer option overall.

In terms of possible future directions, it might be interesting to investigate the feasibility of replacing generative trees with neural networks to handle extremely high-dimensional problems. Another direction would be to instantiate \textit{DeGeTs} framework with alternative methods, such as standard clustering and generative neural networks. Lastly, extending our approach to noisy data sets would likely increase its potential applicability to real world problems.

\section*{Acknowledgments}
All three authors (DM, SS and PC) were supported by the ESRC Research Centre on Micro-Social Change (MiSoC) - ES/S012486/1. This research was supported in part through computational resources provided by the Business and Local Government Data Research Centre BLG DRC (ES/S007156/1) funded by the Economic and Social Research Council (ESRC).

\bibliographystyle{IEEEtranS}
\bibliography{IEEEabrv,main}

\end{document}